\documentclass[11pt,a4paper]{article}
\usepackage[hyperref]{emnlp2018}
\usepackage{times}
\usepackage{latexsym}

\usepackage{url}
\usepackage{amsmath}
\usepackage{amsfonts,amssymb}
\usepackage[T1]{fontenc}
\usepackage{tabularx}
\usepackage{multirow}
\usepackage{booktabs}
\usepackage{subcaption}
\usepackage{float}  
\usepackage{xcolor}

\usepackage{cleveref} 
\usepackage[activate={true,nocompatibility},final,tracking=true,kerning=true,spacing=true,factor=1100,stretch=10,shrink=10]{microtype}
\microtypecontext{spacing=nonfrench}

\newcommand{\mathbold}[1]{\ensuremath{\boldsymbol{\mathbf{#1}}}}

\newcommand{\R}{\mathbb{R}}
\newcommand{\mbh}{\mathbold{h}}
\newcommand{\mbq}{\mathbold{q}}
\newcommand{\mbz}{\mathbold{z}}

\newcommand{\mbH}{\mathbold{H}}

\newcommand{\nateq}{\equiv}
\newcommand{\natind}{\mathbin{\#}}
\newcommand{\natneg}{\mathbin{^{\wedge}}}
\newcommand{\natfor}{\sqsubset}
\newcommand{\natrev}{\sqsupset}
\newcommand{\natalt}{\mathbin{|}}
\newcommand{\natcov}{\mathbin{\smallsmile}}

\usepackage[colorinlistoftodos,prependcaption,textsize=small]{todonotes}

\aclfinalcopy 

\title{The Importance of Being Recurrent \\ for Modeling Hierarchical Structure}

\author{Ke Tran$^1$ \qquad Arianna Bisazza$^2$ \qquad Christof Monz$^1$ \\
  $^{1}$ Informatics Institute, University of Amsterdam\\
  $^{2}$ Leiden Institute of
Advanced Computer Science, Leiden University\\
  {\tt \{m.k.tran,c.monz\}@uva.nl} \qquad {\tt a.bisazza@liacs.leidenuniv.nl}}

\date{}

\begin{document}
\maketitle
\begin{abstract}
  Recent work has shown that recurrent neural networks (RNNs) can implicitly capture and exploit hierarchical information when trained to solve common natural language processing tasks \cite{Blevins:18} such as language modeling \cite{Linzen16,Gulordava:18} and neural machine translation \cite{Shi:2016}.
  In contrast, the ability to model structured data with non-recurrent neural networks has received little attention despite their success in many NLP tasks \citep{Gehring17,Vaswani17}. In this work, we compare the two architectures---\emph{recurrent} versus \emph{non-recurrent}---with respect to their ability to model hierarchical structure and find that recurrency is indeed important for this purpose. The code and data used in our experiments is available at {\small\url{https://github.com/ketranm/fan_vs_rnn}}
\end{abstract}

\section{Introduction}
Recurrent neural networks (RNNs), in particular Long Short-Term Memory networks (LSTMs), have become a dominant tool in natural language processing. While LSTMs appear to be a natural choice for modeling sequential data, recently a class of non-recurrent models \citep{Gehring17,Vaswani17} have
shown competitive performance on sequence modeling. \citet{Gehring17} propose a fully convolutional sequence-to-sequence model that achieves state-of-the-art performance in machine translation. \citet{Vaswani17} introduce Transformer networks that do not use any convolution or recurrent connections while obtaining the best translation performance. These non-recurrent models are appealing due to their highly parallelizable computations on modern GPUs. But do they have the same ability to exploit hierarchical structures \emph{implicitly} in comparison to RNNs?
In this work, we provide a first answer to this question. 

Our interest here is the ability of capturing hierarchical structure without being equipped with explicit structural representations \citep{Bowman15,Tran16,Linzen16}.
We choose Transformer as a non-recurrent model to study in this paper. We refer to Transformer as Fully Attentional Network (FAN) to emphasize this characteristic. Our motivation to favor FANs over convolutional neural networks (CNNs) is that FANs always have full access to the sequence history, making them more suited for modeling long distance dependencies than CNNs. Additionally, FANs promise to be more interpretable than LSTMs by visualizing attention weights.

The rest of the paper is organized as follows: We first highlight the differences between the two architectures (\S\ref{sec:fan_vs_lstm}) and introduce the two tasks (\S\ref{sec:tasks}). Then we provide setup and results for each task (\S\ref{sec:agreement} and \S\ref{sec:logic}) and discuss our findings (\S\ref{sec:conclusion}).

\section{FAN versus LSTM}\label{sec:fan_vs_lstm}
\begin{figure}
    \centering
    \begin{subfigure}[b]{0.2\textwidth}
        \includegraphics[width=\textwidth]{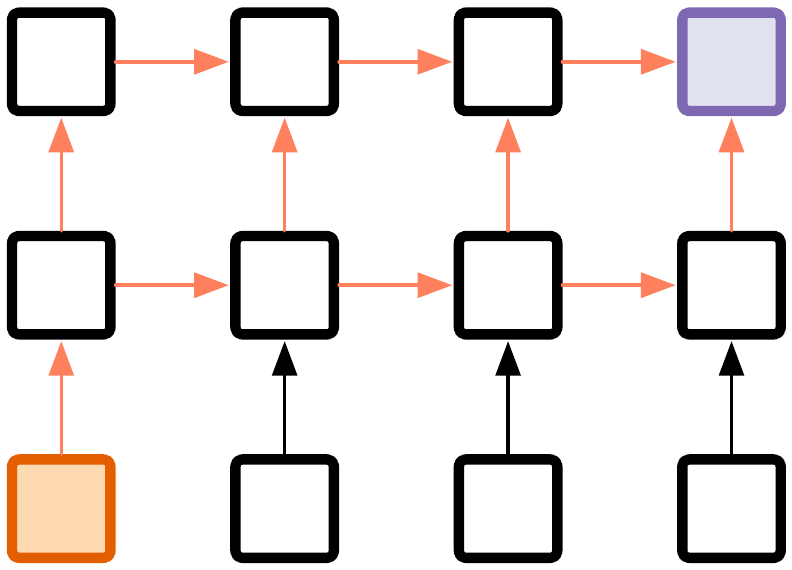}
        \caption{LSTM}
        \label{fig:lstm}
    \end{subfigure}
    \qquad
    \begin{subfigure}[b]{0.2\textwidth}
        \includegraphics[width=\textwidth]{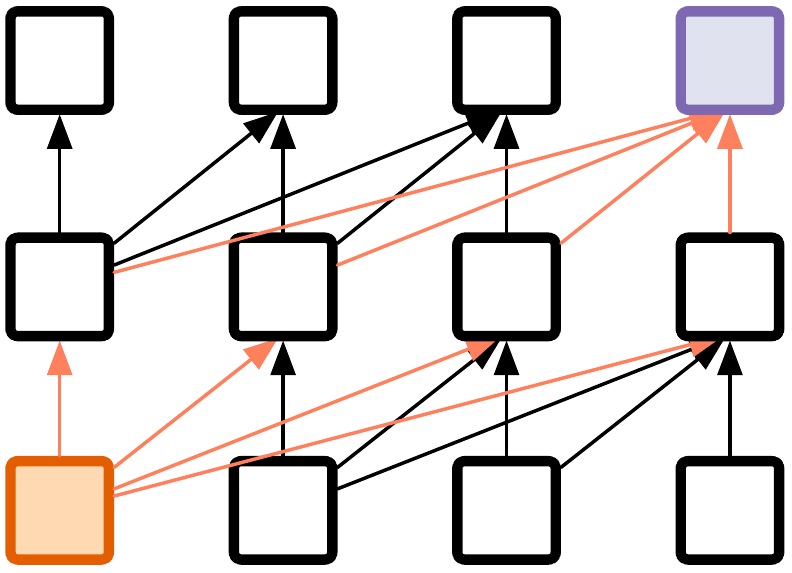}
        \caption{FAN}
        \label{fig:fan}
    \end{subfigure}
    \caption{Diagram showing the main difference between a LSTM and a FAN.  Purple boxes indicate the summarized vector at current time step $t$ which is used to make prediction. Orange arrows indicate the information flow from a previous input to that vector.}\label{fig:archs}
\end{figure}

Conceptually, FANs differ from LSTMs in the way they utilize the previous input to predict the next output. Figure~\ref{fig:archs} depicts the main difference in terms of computation when each model is making predictions. At time step $t$, a FAN can access information from all previous time steps \emph{directly} with $\mathcal{O}(1)$ computational operations. FANs do so by employing a self-attention mechanism to compute the weighted average of all previous input representations. In contrast, LSTMs compress at each time step all  previous information into a single vector \emph{recursively} based on the current input and the previous compressed vector. By their definition, LSTMs require $\mathcal{O}(d)$ computational operations to access the information at time step $t-d$.

For the details of self-attention mechanics in FANs, we refer to the work of \citet{Vaswani17}. We now proceed to measure both models' ability to learn hierarchical structure with a set of controlled experiments.

\section{Tasks}
\label{sec:tasks}
We choose two tasks to study in this work: (1) subject-verb agreement, and (2) logical inference.
The first task was proposed by \citet{Linzen16} to test the ability of recurrent neural networks to capture syntactic dependencies in natural language. The second task was introduced by \citet{Bowman15} to compare tree-based recursive neural networks against sequence-based recurrent networks with respect to their ability to exploit hierarchical structures to make accurate inferences.
The choice of tasks here is important to ensure that both models have to exploit hierarchical structural features \citep{Jia17}.

\section{Subject-Verb Agreement}
\label{sec:agreement}
\citet{Linzen16} propose the task of predicting number agreement between subject and verb in naturally occurring English sentences
as a proxy for the ability of LSTMs to capture hierarchical structure in natural language.
We use the dataset provided by \citet{Linzen16} and follow their experimental protocol of training each model using either (a) a general language model, i.e., next word prediction objective, and (b) an explicit supervision objective, i.e., predicting the number of the verb given its sentence history. Table~\ref{tb:examples} illustrates the training and testing conditions of the task.

\begin{table}[ht]
\caption{Examples of training and test conditions for the two subject-verb agreement subtasks. The full input sentence is ``The  \textbf{keys} to the \underline{cabinet} \textbf{are} on the table'' where verb and subject are bold and intervening nouns are underlined.}
\centering\small
\begin{tabular}{@{}l l l l@{}}
\toprule
 & Input & Train & Test \\
\midrule
(a) & the keys to the cabinet & are & $p$(are) $>$ $p$(is)?\\
(b) &  the keys to the cabinet & plural & plural/singular?\\
\bottomrule
\end{tabular}
\label{tb:examples}
\end{table}%

\noindent \textbf{Data:} Following the original setting, we take 10\% of the data for training, 1\% for validation, and the rest for testing.
The vocabulary consists of the 10k most frequent words, while the remaining words are replaced by their part-of-speech.

\noindent \textbf{Hyperparameters:} 
To allow for a fair comparison, we find the best configuration for each model by running a grid search over the following hyperparameters:
number of layers in $\{2, 3, 4\}$,
dropout rate in $\{0.2, 0.3, 0.5\}$,
embedding size and number of hidden units in $\{128, 256, 512\}$,
number of heads (for FAN) in $\{2, 4\}$,
and learning rate in $\{0.00001, 0.0001, 0.001\}$. The weights of the word embeddings and output layer are shared  \citep{Inan:2016,press:2017}. Models are optimized by Adam \citep{kingma:2014}.

\begin{figure*}[ht]
	\captionsetup[subfigure] {aboveskip=-1pt,belowskip=-1pt}
    \centering
    \begin{subfigure}[b]{0.4\textwidth}
        \includegraphics[width=\textwidth]{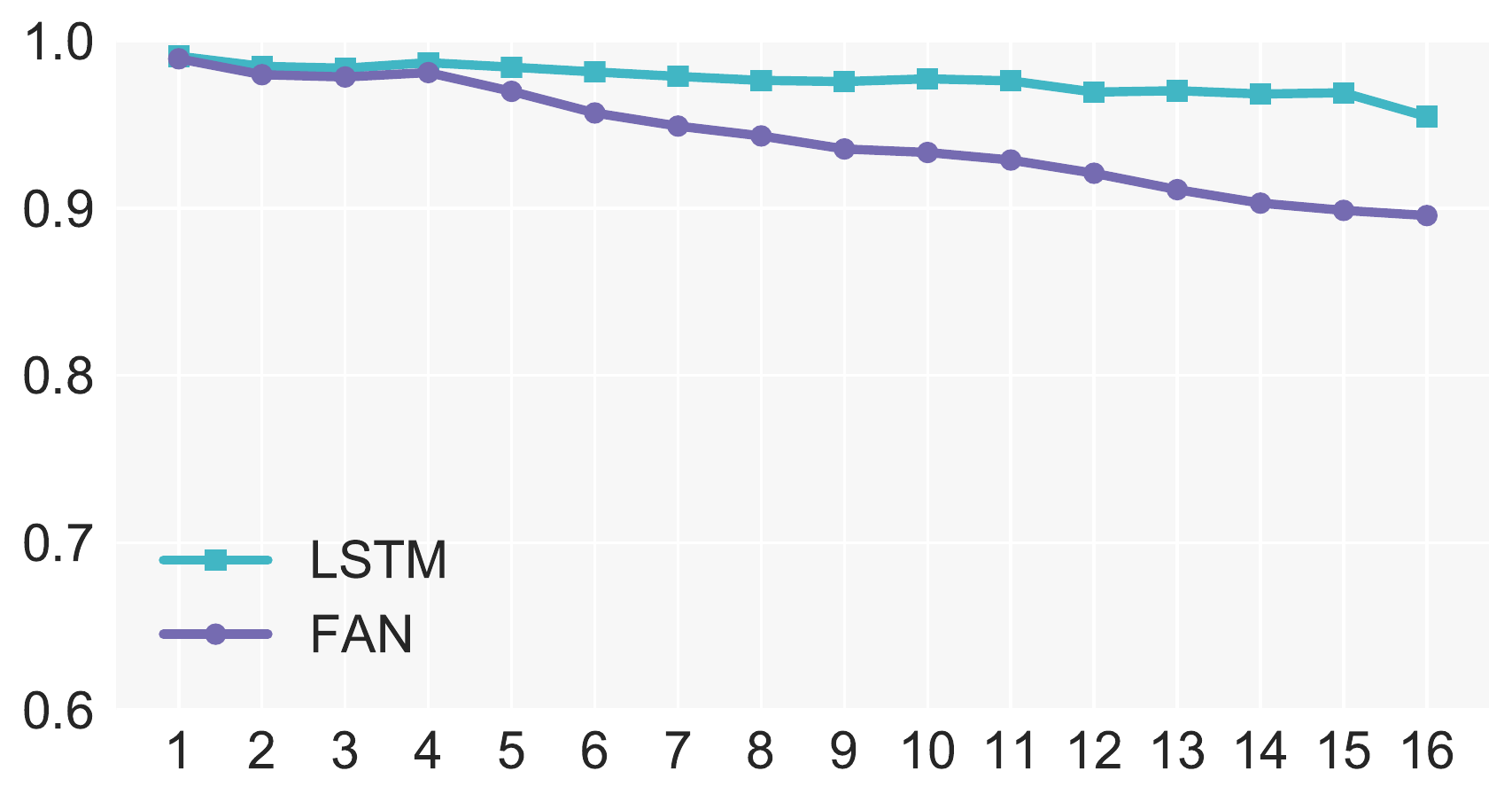}
        \caption{Language model, breakdown by distance}
        \label{fig:distance}
    \end{subfigure}
    \qquad
    \begin{subfigure}[b]{0.4\textwidth}
        \includegraphics[width=\textwidth]{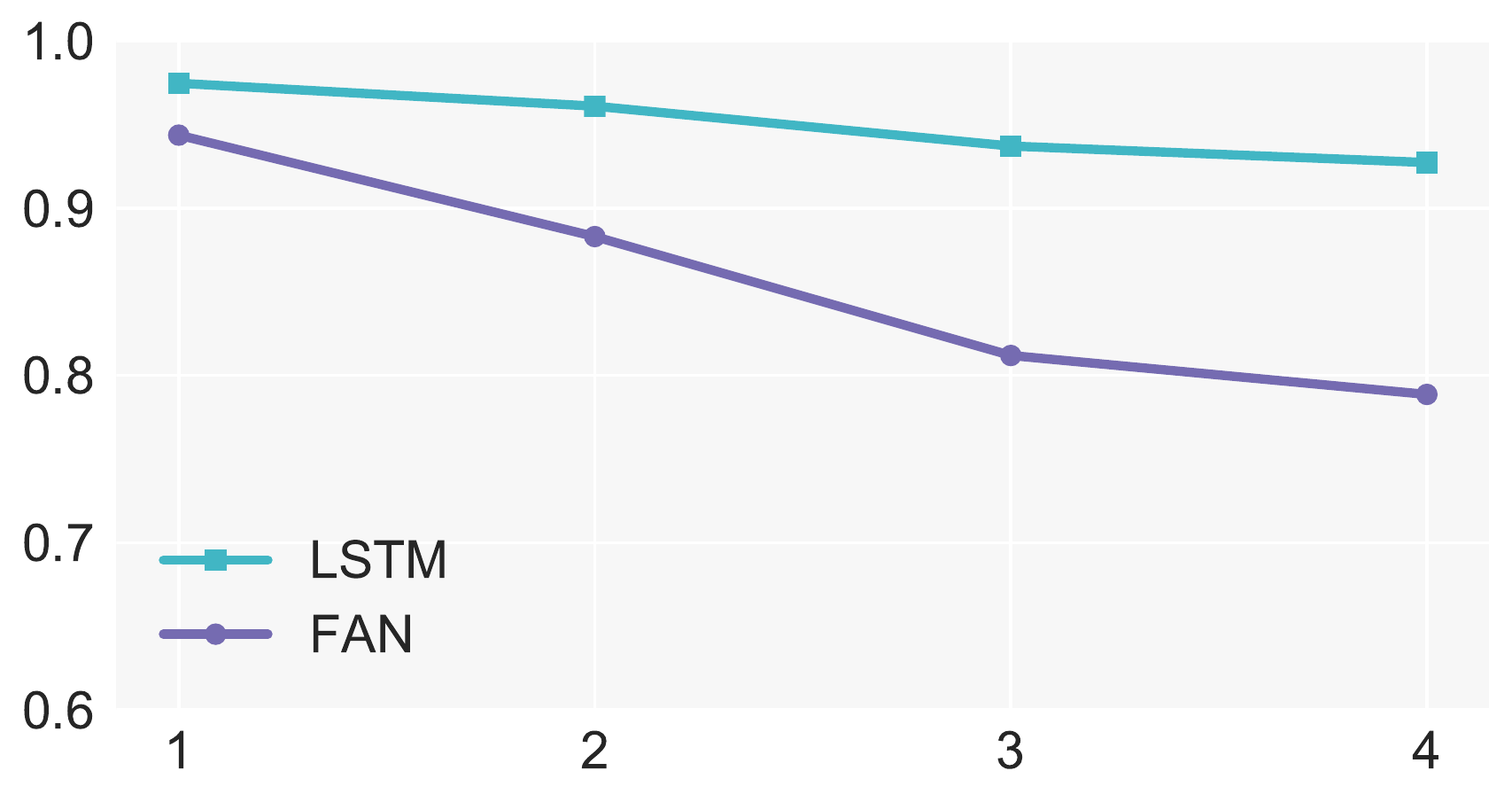}
        \caption{Language model, breakdown by \# attractors}
        \label{fig:intervenings}
    \end{subfigure}
    \qquad
    \begin{subfigure}[b]{0.4\textwidth}
        \includegraphics[width=\textwidth]{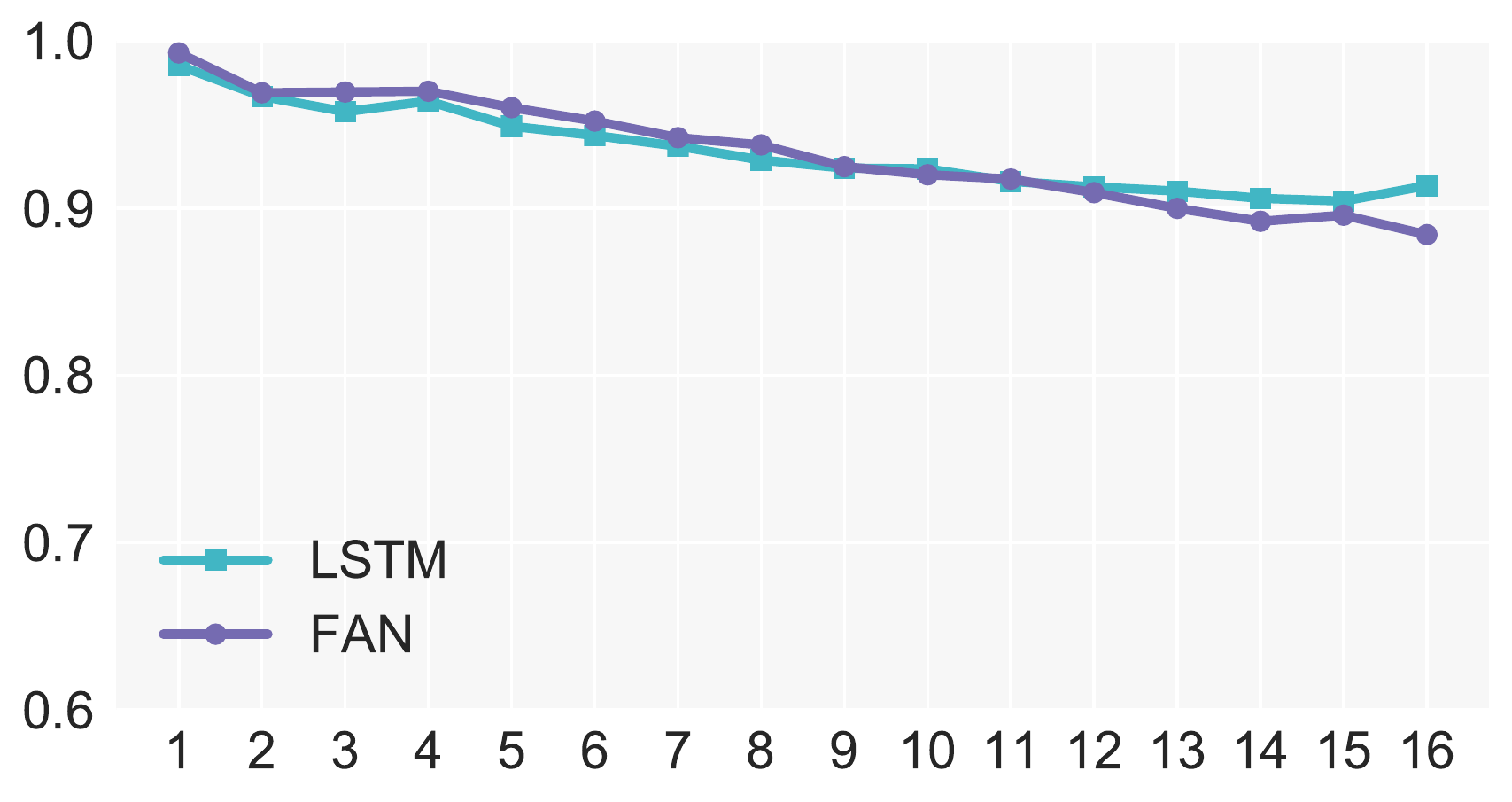}
        \caption{Number prediction, breakdown by distance}
        \label{fig:vp_distance}
    \end{subfigure}
    \qquad
    \begin{subfigure}[b]{0.4\textwidth}
        \includegraphics[width=\textwidth]{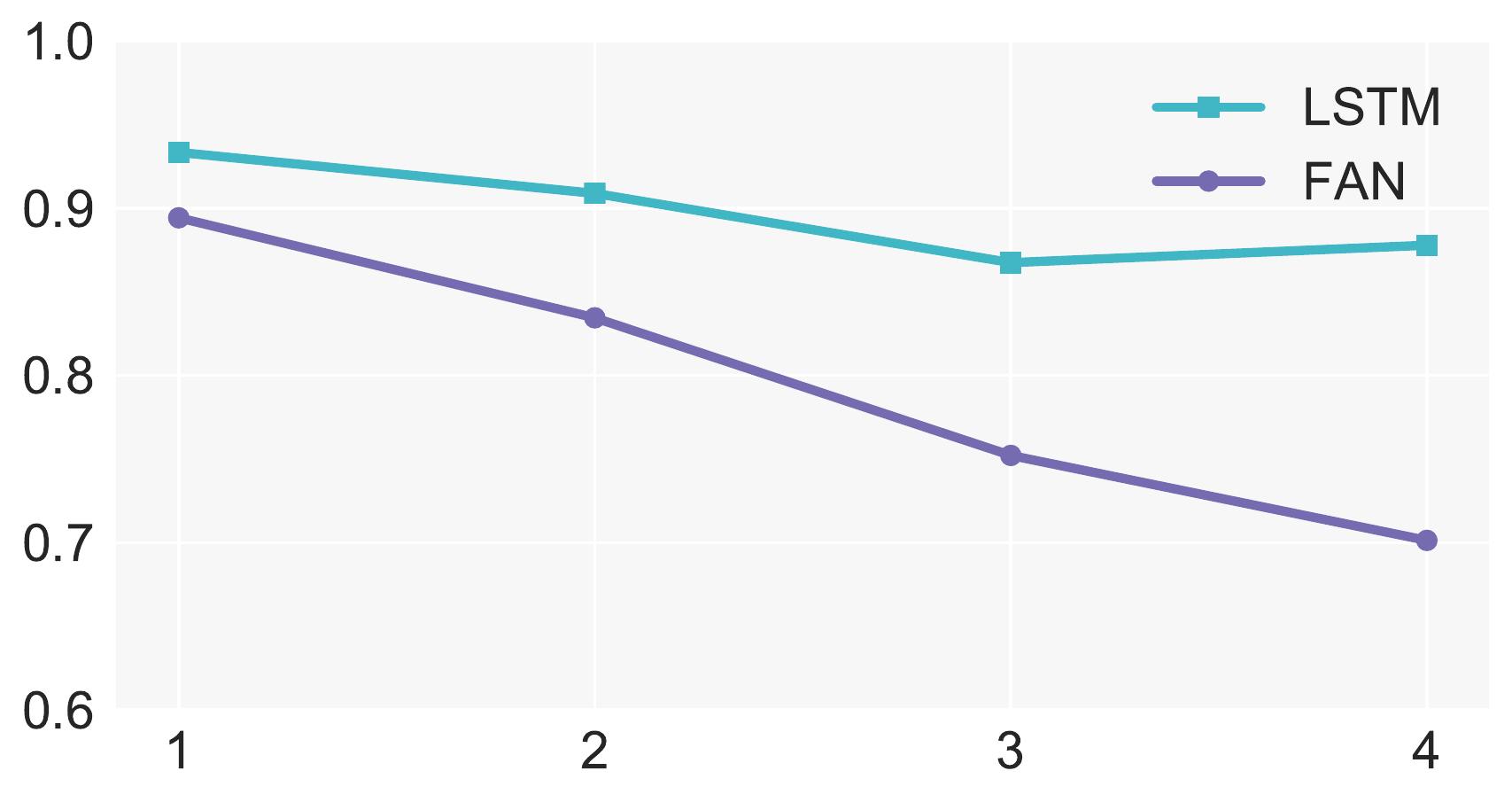}
        \caption{Number prediction, breakdown by \# attractors}
        \label{fig:vp_intervenings}
    \end{subfigure}
    \caption{Results of subject-verb agreement with different training objectives.}
    \label{fig:lmagr}
\end{figure*}

We first assess whether the LSTM and FAN models trained with respect to the language model objective assign higher probabilities to the correctly inflected verbs.
As shown in \Cref{fig:distance,fig:intervenings}, both models achieve high accuracies for this task, but LSTMs consistently outperform FANs.
Moreover, LSTMs are clearly more robust than FANs with respect to task difficulty, measured both in terms of word distance and number of agreement attractors\footnote{Agreement attractors are intervening nouns with the opposite number from the subject.} between subject and verb.
\citet{Christiansen2016,Cornish2017} have argued that human memory limitations give rise to important characteristics of natural language, including its hierarchical structure.
Similarly, our experiments suggest that, by compressing the history into a single vector before making predictions, LSTMs are forced to better learn the input structure.
On the other hand, despite having direct access to all words in their history, FANs are less capable of detecting the verb's subject.
We note that the validation perplexities of the LSTM and FAN are 67.06 and  69.14, respectively.

Secondly, we evaluate FAN and LSTM models explicitly trained to predict the verb number (\Cref{fig:vp_distance,fig:vp_intervenings}).
Again, we observe that LSTMs consistently outperform FANs. This is a particularly interesting result since the self-attention mechanism in FANs connects two words in any position with a $\mathcal{O}(1)$ number of executed operations, whereas RNNs require more recurrent operations. Despite this apparent advantage of FANs, the performance gap between FANs and LSTMs increases with the distance and number of attractors.%
\footnote{We note that our LSTM results are better than those in \citet{Linzen16}. Also surprising is that the language model objective yields higher accuracies than the number prediction objective. 
We believe this may be due to better model optimization and to the embedding-output layer weight sharing, but we leave a thorough investigation to future work.}

\begin{figure}
  \includegraphics[width=0.47\textwidth]{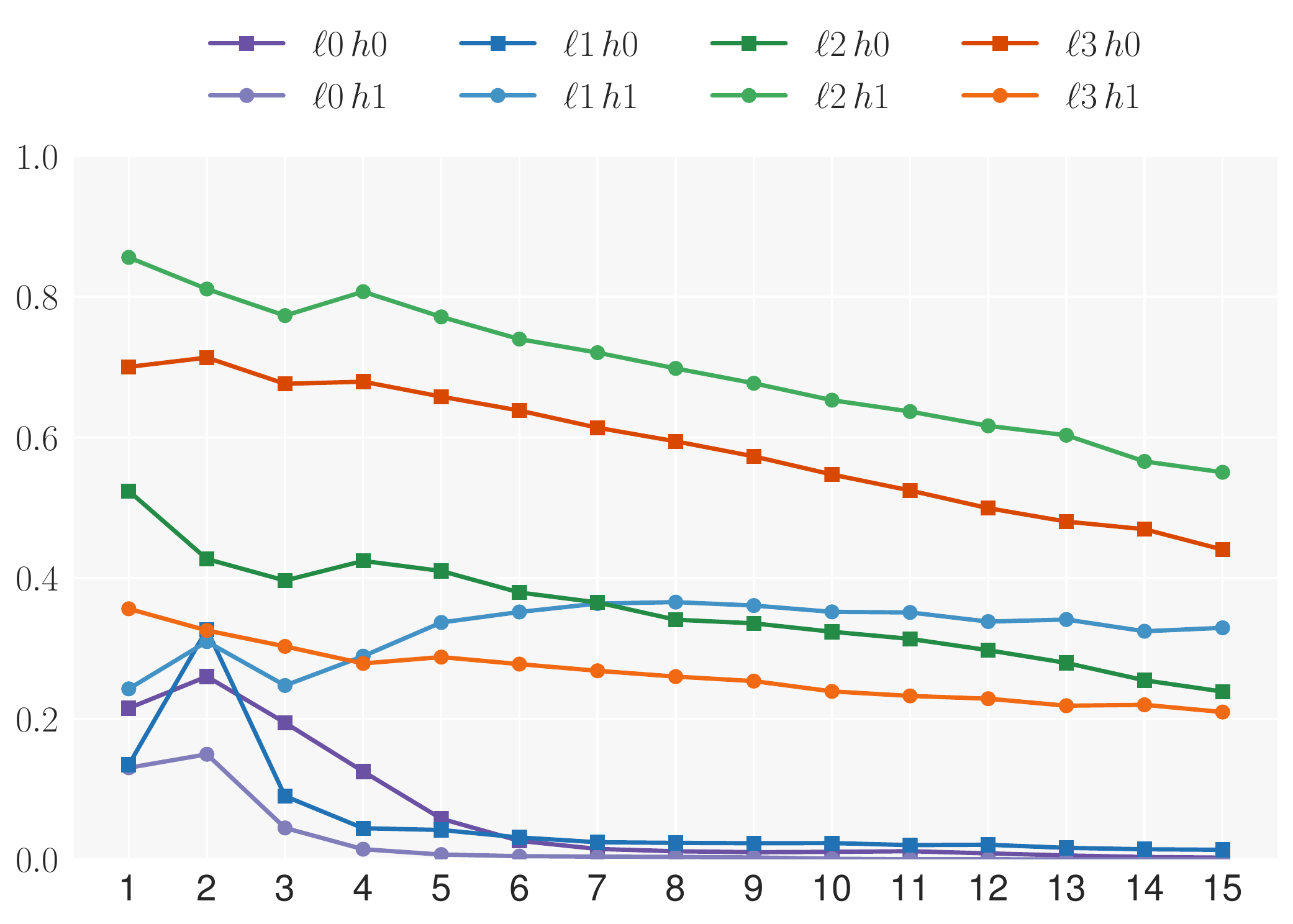}
  \caption{Proportion of times the subject is the most attended word by different heads at different layers ($\ell3$ is the highest layer).
  Only cases where the model made a correct prediction are shown.}
  \label{fig:attn_vp_distance}
\end{figure}

To gain further insights into our results, we examine the attention weights computed by FANs during verb-number prediction (supervised objective).
Specifically, for each attention head at each layer of the FAN, we compute the percentage of times the subject is the most attended word among all words in the history.
Figure~\ref{fig:attn_vp_distance} shows the results for all cases where the model made the correct prediction.
While it is hard to interpret the exact role of attention for different heads and at different layers,
we find that some of the attention heads at the higher layers ($\ell2$ $h1$, $\ell3$ $h0$) frequently point to the subject with an accuracy that decreases linearly with the distance between subject and verb.

\section{Logical inference}\label{sec:logic}
In this task, we choose the artificial language introduced by \citet{Bowman15}.
The vocabulary of this language includes six word types \{\textit{a, b, c, d, e, f}\} and three logical operators \{\textit{or}, \textit{and}, \textit{not}\}.
The task consists of predicting one of seven mutually exclusive logical relations that describe the relationship between a pair of sentences: entailment ($\natfor$, $\natrev$), equivalence ($\nateq$), exhaustive and non-exhaustive contradiction ($\natneg$, $\natalt$), and two types of semantic independence ($\natind$, $\natcov$).
We generate 60,000 samples\footnote{\url{https://github.com/sleepinyourhat/vector-entailment}} with the number of logical operations ranging from 1 to 12. The train/dev/test dataset ratios are set to 0.8/0.1/0.1. Here are some samples of the training data:
 \begin{table}[htp]
   \begin{center}\small
   \begin{tabular}{@{} r @{\ } c @{\ } l}
   ( d ( or f ) ) & $\natrev$ & ( f ( and a ) ) \\
   ( d ( and ( c ( or d ) ) ) )  & $\natind$ &  ( not f ) \\
   ( not ( d ( or ( f ( or c ) ) ) ) )  & $\natfor$  &   ( not ( c ( and ( not d ) ) ) ) \\
   \end{tabular}
   \end{center}
   \label{tb:lg_samples}
 \end{table}%

\noindent\textbf{Why artificial data?} Despite the simplicity of the language, this task is not trivial.
To correctly classify logical relations, the model must learn nested structures as well as the scope of logical operators.
We verify the difficulty of the task by training three bag-of-words models followed by sum/average/max-pooling.
The best of the three models
achieve less than 59\% accuracy on the logical inference versus 77\% on the Stanford Natural Language Inference (SNLI) corpus \citep{Bowman:2015:EMNLP}. This shows that the SNLI task can be largely solved by exploiting  shallow features without understanding the underlying linguistic structures, which has also been pointed out by recent work \citep{glockner_acl18,Gururangan:18}.

Concurrently to our work \citet{evan:18} proposed an alternative data set for logical inference and also found that a FAN model underperformed various other architectures including LSTMs.

\subsection{Models}
We follow the general architecture proposed in \cite{Bowman15}: Premise and hypothesis sentences are encoded by fixed-size vectors. These two vectors are then concatenated and fed to a 3-layer feed-forward neural network with ReLU nonlinearities to perform 7-way classification of the logical relation.

The LSTM architecture used in this experiment is similar to that of \citet{Bowman15}. We simply take the last hidden state of the top LSTM layer as a fixed-size vector representation of the sentence. Here, we use a 2-layer LSTM with skip connections.
The FAN maps a sentence $x$ of length $n$ to $\mbH = [\mbh_1, \dots, \mbh_n] \in \R^{d\times n}$.
To obtain a fixed-size representation $\mbz$, we use a self-attention layer with two trainable queries $\mbq_1, \mbq_2 \in \R^{1\times d}$:
\begin{align}
\mbz_i &= \textrm{softmax}\left(\frac{\mbq_i \mbH}{\sqrt d}\right) \mbH^\top \quad i\in\{1, 2\} \nonumber\\
\mbz &= [\mbz_1, \mbz_2] \nonumber
\end{align}

\noindent
We find the best hyperparameters for each model by running a grid search as explained in \S\ref{sec:agreement}.

\subsection{Results}
\begin{figure}
	\captionsetup[subfigure]{aboveskip=-1pt,belowskip=-1pt}
    \centering
    \begin{subfigure}[b]{0.4\textwidth}
        \includegraphics[width=\textwidth]{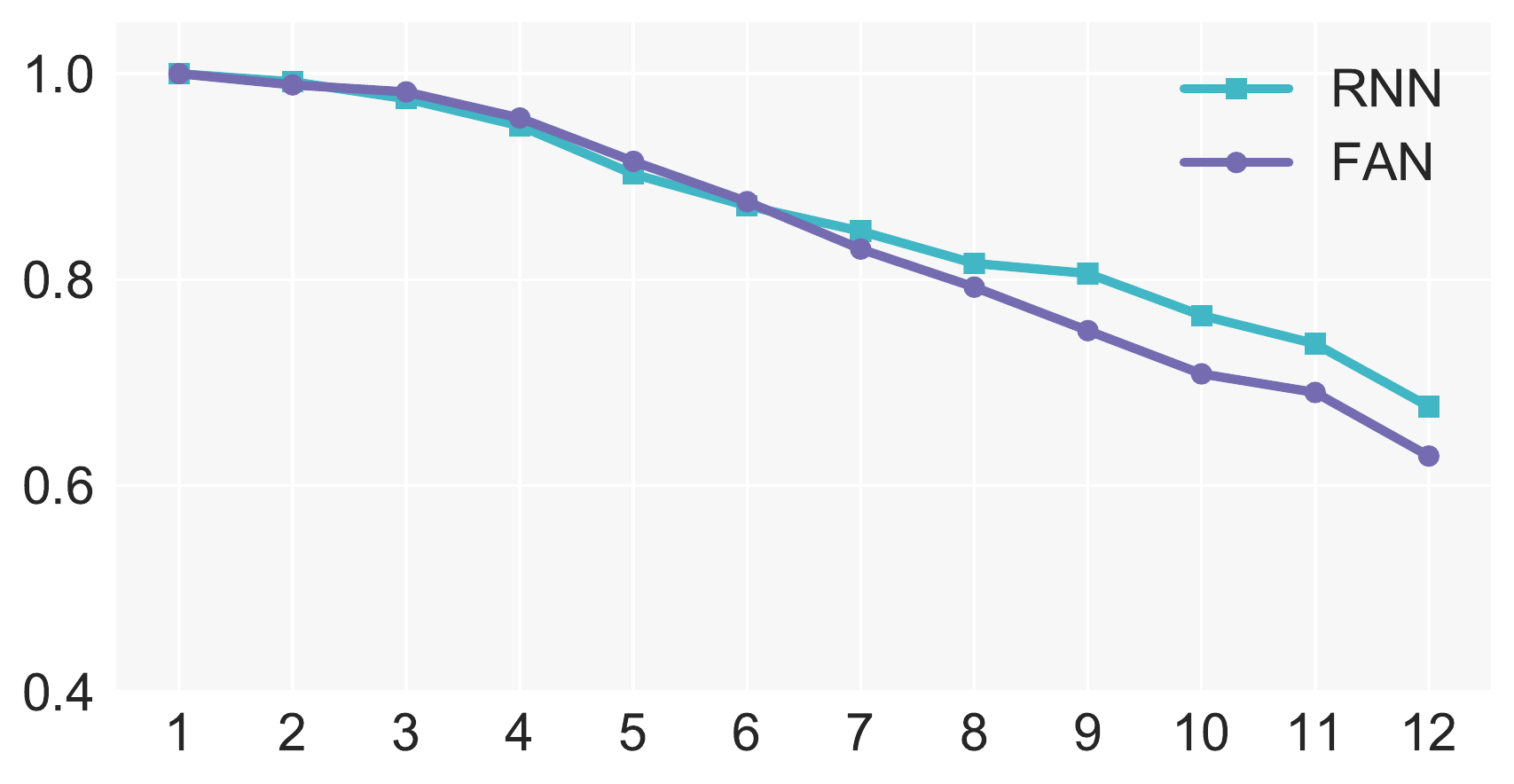}
        \caption{$n\le 12$}
        \label{fig:plbin12}
    \end{subfigure}
    \qquad
    \begin{subfigure}[b]{0.4\textwidth}
        \includegraphics[width=\textwidth]{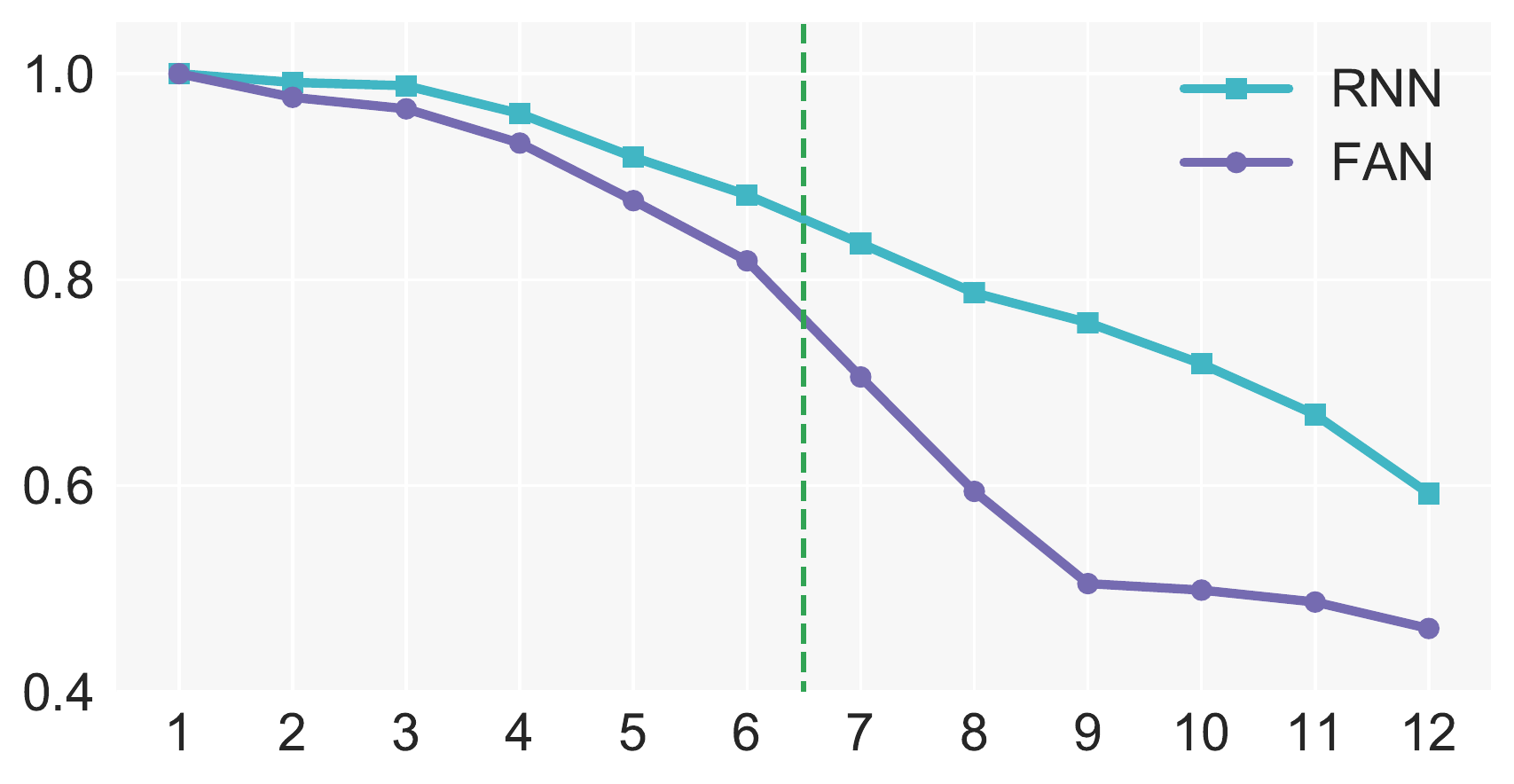}
        \caption{$n\le 6$}
        \label{fig:plbin6}
    \end{subfigure}
    \caption{Results of logical inference when training on all data (a) or only on samples with at most $n$ logical operators (b).}
    \label{fig:plresults}
\end{figure}
Following the experimental protocol of \citet{Bowman15}, the data is divided into 13 bins based on the number of logical operators. Both FANs and LSTMs are trained on samples with at most $n$ logical operators and tested on all bins.
Figure~\ref{fig:plresults} shows the result of the experiments with $n\le6$ and $n\le12$. We see that FANs and LSTMs perform similarly when trained on the whole dataset (Figure~\ref{fig:plbin12}). However when trained on a subset of the data (Figure~\ref{fig:plbin6}), LSTMs obtain better accuracies on similar examples ($n\le6$) and generalize better on longer examples ($6<n\le 12$).

\section{Discussion and Conclusion}
\label{sec:conclusion}
We have compared a recurrent architecture (LSTM) to a non-recurrent one (FAN) with respect to the ability of capturing the underlying hierarchical structure of sequential data. Our experiments show that LSTMs slightly but consistently outperform FANs. We found that LSTMs are notably more robust with respect to the presence of misleading features in the agreement task, whether trained with explicit supervision or with a general language model objective. Secondly, we found that LSTMs generalize better than FANs to longer sequences in a logical inference task.
These findings suggest that recurrency is a key model property which should not be sacrificed for efficiency when hierarchical structure matters for the task.

This does not imply that LSTMs should always be preferred over non-recurrent architectures. In fact, both FAN- and CNN-based networks have proved to perform comparably or better than LSTM-based ones on a very complex task like machine translation \cite{Gehring17,Vaswani17}.
Nevertheless, we believe that the ability of capturing hierarchical information in sequential data remains a fundamental need for building intelligent systems that can understand and process language.
Thus we hope that our insights will be useful towards building the next generation of neural networks.

\section*{Acknowledgments}
This research was funded in part by the Netherlands Organization for Scientific Research (NWO) under project numbers 639.022.213, 612.001.218, and 639.021.646.

\bibliography{emnlp2018}

\begin{thebibliography}{18}
\expandafter\ifx\csname natexlab\endcsname\relax\def\natexlab#1{#1}\fi

\bibitem[{Blevins et~al.(2018)Blevins, Levy, and Zettlemoyer}]{Blevins:18}
Terra Blevins, Omer Levy, and Luke Zettlemoyer. 2018.
\newblock Deep rnns encode soft hierarchical syntax.
\newblock In \emph{Proceedings of the 56th Annual Meeting of the Association
  for Computational Linguistics (Volume 2: Short Papers)}, pages 14--19,
  Melbourne, Australia. Association for Computational Linguistics.

\bibitem[{Bowman et~al.(2015{\natexlab{a}})Bowman, Angeli, Potts, and
  Manning}]{Bowman:2015:EMNLP}
Samuel~R. Bowman, Gabor Angeli, Christopher Potts, and Christopher~D. Manning.
  2015{\natexlab{a}}.
\newblock A large annotated corpus for learning natural language inference.
\newblock In \emph{Proceedings of the 2015 Conference on Empirical Methods in
  Natural Language Processing}, pages 632--642, Lisbon, Portugal. Association
  for Computational Linguistics.

\bibitem[{Bowman et~al.(2015{\natexlab{b}})Bowman, Manning, and
  Potts}]{Bowman15}
Samuel~R. Bowman, Christopher~D. Manning, and Christopher Potts.
  2015{\natexlab{b}}.
\newblock Tree-structured composition in neural networks without
  tree-structured architectures.
\newblock In \emph{Proceedings of the 2015th International Conference on
  Cognitive Computation: Integrating Neural and Symbolic Approaches - Volume
  1583}, COCO'15, pages 37--42.

\bibitem[{Christiansen and Chater(2016)}]{Christiansen2016}
Morten~H. Christiansen and Nick Chater. 2016.
\newblock The now-or-never bottleneck: A fundamental constraint on language.
\newblock \emph{Behavioral and Brain Sciences}, 39.

\bibitem[{Cornish et~al.(2017)Cornish, Dale, Kirby, and
  Christiansen}]{Cornish2017}
Hannah Cornish, Rick Dale, Simon Kirby, and Morten~H Christiansen. 2017.
\newblock Sequence memory constraints give rise to language-like structure
  through iterated learning.
\newblock \emph{PloS one}, 12(1):e0168532.

\bibitem[{Evans et~al.(2018)Evans, Saxton, Amos, Kohli, and
  Grefenstette}]{evan:18}
Richard Evans, David Saxton, David Amos, Pushmeet Kohli, and Edward
  Grefenstette. 2018.
\newblock Can neural networks understand logical entailment?
\newblock In \emph{ICLR}.

\bibitem[{Gehring et~al.(2017)Gehring, Auli, Grangier, Yarats, and
  Dauphin}]{Gehring17}
Jonas Gehring, Michael Auli, David Grangier, Denis Yarats, and Yann~N. Dauphin.
  2017.
\newblock Convolutional sequence to sequence learning.
\newblock In \emph{Proceedings of the 34th International Conference on Machine
  Learning}, volume~70 of \emph{Proceedings of Machine Learning Research},
  pages 1243--1252, International Convention Centre, Sydney, Australia. PMLR.

\bibitem[{Glockner et~al.(2018)Glockner, Shwartz, and
  Goldberg}]{glockner_acl18}
Max Glockner, Vered Shwartz, and Yoav Goldberg. 2018.
\newblock Breaking nli systems with sentences that require simple lexical
  inferences.
\newblock In \emph{The 56th Annual Meeting of the Association for Computational
  Linguistics (ACL)}, Melbourne, Australia.

\bibitem[{Gulordava et~al.(2018)Gulordava, Bojanowski, Grave, Linzen, and
  Baroni}]{Gulordava:18}
Kristina Gulordava, Piotr Bojanowski, Edouard Grave, Tal Linzen, and Marco
  Baroni. 2018.
\newblock Colorless green recurrent networks dream hierarchically.
\newblock In \emph{Proceedings of the 2018 Conference of the North American
  Chapter of the Association for Computational Linguistics: Human Language
  Technologies, Volume 1 (Long Papers)}, pages 1195--1205. Association for
  Computational Linguistics.

\bibitem[{Gururangan et~al.(2018)Gururangan, Swayamdipta, Levy, Schwartz,
  Bowman, and Smith}]{Gururangan:18}
Suchin Gururangan, Swabha Swayamdipta, Omer Levy, Roy Schwartz, Samuel Bowman,
  and Noah~A. Smith. 2018.
\newblock Annotation artifacts in natural language inference data.
\newblock In \emph{Proceedings of the 2018 Conference of the North American
  Chapter of the Association for Computational Linguistics: Human Language
  Technologies, Volume 2 (Short Papers)}, pages 107--112, New Orleans,
  Louisiana. Association for Computational Linguistics.

\bibitem[{Inan et~al.(2017)Inan, Khosravi, and Socher}]{Inan:2016}
Hakan Inan, Khashayar Khosravi, and Richard Socher. 2017.
\newblock Tying word vectors and word classifiers: A loss framework for
  language modeling.
\newblock \emph{ICLR}.

\bibitem[{Jia and Liang(2017)}]{Jia17}
Robin Jia and Percy Liang. 2017.
\newblock Adversarial examples for evaluating reading comprehension systems.
\newblock In \emph{Proceedings of the 2017 Conference on Empirical Methods in
  Natural Language Processing}, pages 2021--2031, Copenhagen, Denmark.
  Association for Computational Linguistics.

\bibitem[{Kingma and Ba(2015)}]{kingma:2014}
Diederik Kingma and Jimmy Ba. 2015.
\newblock Adam: A method for stochastic optimization.
\newblock In \emph{Proceedings of ICLR}, San Diego, CA, USA.

\bibitem[{Linzen et~al.(2016)Linzen, Dupoux, and Goldberg}]{Linzen16}
Tal Linzen, Emmanuel Dupoux, and Yoav Goldberg. 2016.
\newblock Assessing the ability of lstms to learn syntax-sensitive
  dependencies.
\newblock \emph{Transactions of the Association for Computational Linguistics},
  4:521--535.

\bibitem[{Press and Wolf(2017)}]{press:2017}
Ofir Press and Lior Wolf. 2017.
\newblock Using the output embedding to improve language models.
\newblock In \emph{Proceedings of the 15th Conference of the European Chapter
  of the Association for Computational Linguistics: Volume 2, Short Papers},
  pages 157--163. Association for Computational Linguistics.

\bibitem[{Shi et~al.(2016)Shi, Padhi, and Knight}]{Shi:2016}
Xing Shi, Inkit Padhi, and Kevin Knight. 2016.
\newblock Does string-based neural mt learn source syntax?
\newblock In \emph{Proceedings of the 2016 Conference on Empirical Methods in
  Natural Language Processing}, pages 1526--1534, Austin, Texas. Association
  for Computational Linguistics.

\bibitem[{Tran et~al.(2016)Tran, Bisazza, and Monz}]{Tran16}
Ke~Tran, Arianna Bisazza, and Christof Monz. 2016.
\newblock Recurrent memory networks for language modeling.
\newblock In \emph{Proceedings of the 2016 Conference of the North American
  Chapter of the Association for Computational Linguistics: Human Language
  Technologies}, pages 321--331, San Diego, California. Association for
  Computational Linguistics.

\bibitem[{Vaswani et~al.(2017)Vaswani, Shazeer, Parmar, Uszkoreit, Jones,
  Gomez, Kaiser, and Polosukhin}]{Vaswani17}
Ashish Vaswani, Noam Shazeer, Niki Parmar, Jakob Uszkoreit, Llion Jones,
  Aidan~N Gomez, \L~ukasz Kaiser, and Illia Polosukhin. 2017.
\newblock Attention is all you need.
\newblock In I.~Guyon, U.~V. Luxburg, S.~Bengio, H.~Wallach, R.~Fergus,
  S.~Vishwanathan, and R.~Garnett, editors, \emph{Advances in Neural
  Information Processing Systems 30}, pages 6000--6010. Curran Associates, Inc.

\end{thebibliography}
\bibliographystyle{acl_natbib_nourl}
\end{document}